\documentclass{article}

\usepackage[preprint]{neurips_2023}
\usepackage[T1]{fontenc}    
\usepackage[utf8]{inputenc} 
\usepackage{algorithm, algorithmicx, algpseudocode}
\usepackage{amsmath, amsfonts, amssymb}
\usepackage{array}
\usepackage{booktabs}
\usepackage{collcell}
\usepackage{datatool}
\usepackage{enumitem}
\usepackage{environ}
\usepackage{etoolbox}
\usepackage{flushend}
\usepackage{graphicx}
\usepackage{lipsum}
\usepackage{microtype}
\usepackage{multirow}
\usepackage{nicefrac}
\usepackage{printlen}
\usepackage{soul}
\usepackage{svg}
\usepackage{url}
\usepackage{wrapfig}
\usepackage{xcolor,colortbl}
\usepackage{xstring}

\definecolor{Gray}{gray}{0.85}
\definecolor{cite}{HTML}{53769A}
\definecolor{ref}{HTML}{13BF28}

\definecolor{lightcornflowerblue}{rgb}{0.6, 0.81, 0.93}
\definecolor{lightkhaki}{rgb}{0.94, 0.9, 0.55}
\definecolor{lightmauve}{rgb}{0.86, 0.82, 1.0}
\definecolor{lightgreen}{rgb}{0.56, 0.93, 0.56}

\definecolor{PatternA}{RGB}{180, 22,  0   }
\definecolor{PatternC}{RGB}{23,  77,  127 }
\definecolor{PatternB}{RGB}{55,  144, 48  }

\usepackage[colorlinks=true, citecolor=cite, linkcolor=ref]{hyperref}

\newcolumntype{H}{>{\setbox0=\hbox\bgroup}c<{\egroup}@{}}

\newcommand{\augment}{dynamic scanning augmentation}

\newcommand{\com}[1]{\iffalse~#1~\fi}%

\algrenewcommand{\algorithmiccomment}[1]{\hfill$\blacktriangleright$ #1}

\newcolumntype{X}{>{\columncolor{lightcornflowerblue}}c}
\newcolumntype{Y}{>{\columncolor{lightkhaki}}c}
\newcolumntype{Z}{>{\columncolor{lightmauve}}c}
\newcolumntype{P}{>{\columncolor{lightgreen}}c}

\newcommand{\noimage}{%
  \setlength{\fboxsep}{-\fboxrule}%
  \fbox{\phantom{\rule{100pt}{100pt}}File missing\phantom{\rule{100pt}{100pt}}}
}
\let\includegraphicsoriginal\includegraphics%
\renewcommand{\includegraphics}[2][width=\textwidth]{\IfFileExists{#2}{\includegraphicsoriginal[#1]{#2}}{\noimage}}

\emergencystretch=\maxdimen%
\everypar{\looseness=-1 }

\newcounter{descriptcount}

\newcounter{CurrentRow}
\newcounter{CurrentColumn}
\newtoggle{DoneWithFirstRow}

\newcommand*{\FirstColumn}[1]{%
    \IfEq{\arabic{CurrentColumn}}{0}{%
        \global\togglefalse{DoneWithFirstRow}%
        \setcounter{CurrentRow}{1}
    }{%
        \global\toggletrue{DoneWithFirstRow}%
        \stepcounter{CurrentRow}%
    }%
    \setcounter{CurrentColumn}{0}%
    \NewData{#1}%
}
\newcommand*{\NewData}[1]{%
    \dtlexpandnewvalue%
    \stepcounter{CurrentColumn}%
    \iftoggle{DoneWithFirstRow}{%
        \dtlgetrow{TransposedTabularDB}{\arabic{CurrentColumn}}%
        \dtlappendentrytocurrentrow{\Alph{CurrentRow}}{#1}%
        \dtlrecombine%
    }{%
        \DTLnewrow{TransposedTabularDB}%
        \DTLnewdbentry{TransposedTabularDB}{\Alph{CurrentRow}}{#1}%
    }%
}%

\newcolumntype{+}{>{\global\let\currentrowstyle\relax}}
\newcolumntype{^}{>{\currentrowstyle}}
\newcolumntype{F}{>{\collectcell\FirstColumn}c<{\endcollectcell}}
\newcolumntype{C}{>{\collectcell\NewData}{c}<{\endcollectcell}}

\newtoggle{EncounteredDataRow}

\newsavebox{\TempBox}
\DTLnewdb{TransposedTabularDB}

\NewEnviron{Ttabular}[1]{%
    \setcounter{CurrentColumn}{0}%
    \global\togglefalse{EncounteredDataRow}%
    \savebox{\TempBox}{%
        \begin{tabular}{FCCCCCC}
            \BODY%
        \end{tabular}%
    }%
    \begin{tabular}{#1}\toprule%
    \DTLforeach*{TransposedTabularDB}{\Aa=A, \Ba=B, \Ca=C}{%
      \DTLiffirstrow{}{\\\midrule}%
        \Aa&\Ba&\Ca%
    }\\\bottomrule%
    \end{tabular}%
}%

\def\Tableref#1{Table~\ref{#1}}

\def\Twofigref#1#2{Figures~\ref{#1}~and~\ref{#2}}

\def\eqref#1{equation~\ref{#1}}

\usepackage{mathabx,graphicx}

\def\Figref#1{Figure~\ref{#1}}

\def\ceil#1{\lceil~#1~\rceil}

\def\1{\bm{1}}

\DeclareMathAlphabet{\mathsfit}{\encodingdefault}{\sfdefault}{m}{sl}
\SetMathAlphabet{\mathsfit}{bold}{\encodingdefault}{\sfdefault}{bx}{n}

\title{Improving Robustness for Vision Transformer with a Simple Dynamic Scanning Augmentation}

\author{%
  Shashank Kotyan \quad Danilo Vasconcellos Vargas \\
  Kyushu University \\
  \texttt{kotyan.shashank.651@s.kyushu-u.ac.jp} \\
}

\begin{document}

\maketitle

\begin{abstract}
Vision Transformer (ViT) has demonstrated promising performance in computer vision tasks, comparable to state-of-the-art neural networks.
Yet, this new type of deep neural network architecture is vulnerable to adversarial attacks limiting its capabilities in terms of robustness. 
This article presents a novel contribution aimed at further improving the accuracy and robustness of ViT, particularly in the face of adversarial attacks.
We propose an augmentation technique called `Dynamic Scanning Augmentation' that leverages dynamic input sequences to adaptively focus on different patches, thereby maintaining performance and robustness.
Our detailed investigations reveal that this adaptability to the input sequence induces significant changes in the attention mechanism of ViT, even for the same image.
We introduce four variations of Dynamic Scanning Augmentation, outperforming ViT in terms of both robustness to adversarial attacks and accuracy against natural images, with one variant showing comparable results.
By integrating our augmentation technique, we observe a substantial increase in ViT's robustness, improving it from $17\%$ to $92\%$ measured across different types of adversarial attacks.
These findings, together with other comprehensive tests, indicate that Dynamic Scanning Augmentation enhances accuracy and robustness by promoting a more adaptive type of attention.
In conclusion, this work contributes to the ongoing research on Vision Transformers by introducing Dynamic Scanning Augmentation as a technique for improving the accuracy and robustness of ViT.
The observed results highlight the potential of this approach in advancing computer vision tasks and merit further exploration in future studies.

\end{abstract}

\section{Introduction}

Transformers \cite{vaswani2017attention}, a type of neural network based on self-attention, has become the standard for many Text-based NLP tasks.
Recently, many neural network architectures based on transformers and self-attention have been proposed for vision tasks that outperform other architectures. \cite{dosovitskiy2020image,ramachandran2019stand,wu2020visual,liu2021swin,tang2021sensory}.

Although Vision Transformer (ViT) seems promising for the vision tasks in terms of accuracy, it suffers from a few underlying problems like,

\noindent
a) Robustness against adversarial attacks \cite{shao2022adversarial,mahmood2021robustness}, and

\noindent
b) Extraction of local features \cite{han2021transformer,yuan2021tokens}.

Therefore, the potential of self-attention and transformers is yet to be fully utilized, especially regarding solving vision problems.
Until now, augmentations have focused on modifying the image pixels based on transformations since the whole image is processed at once.
However, the definition of augmentation can be extended to input sequences and how these input sequences are created for the transformers.
Transformers process images as a sequence of patches, allowing for a new kind of augmentation based on the dynamic scanning of images.

\begin{figure*}[!t]
    \centering
    \includegraphics[width=\textwidth]{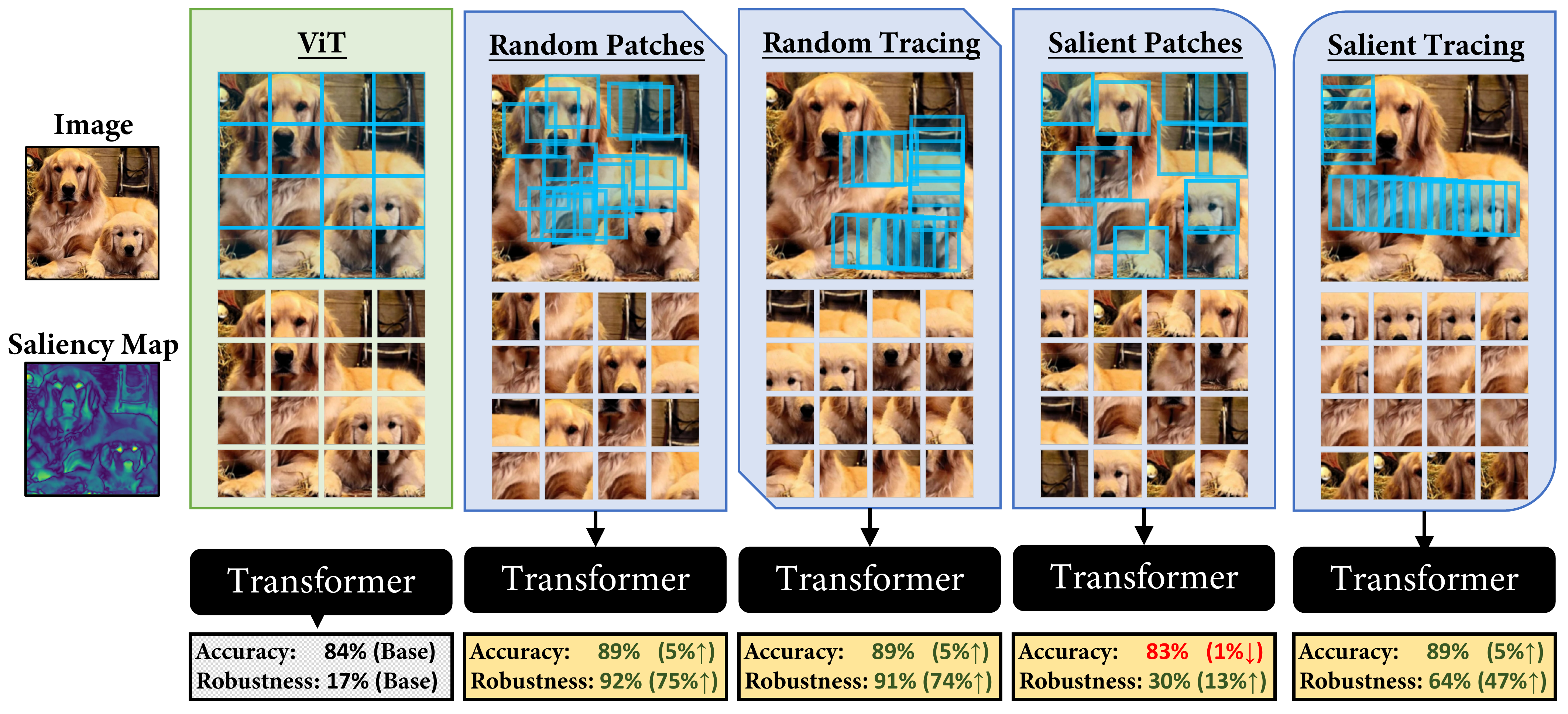}
    \caption{
        Overview of the different variations of proposed \augment~ and their performance on the CIFAR-10 dataset.
        We report the accuracy over non-attacked images and mean robust accuracy over three different black-box adversarial attacks.
        All of the transformers employed with \augment~ substantially improve the performance and robustness of ViT.
    }
    \label{method}
\end{figure*}

We propose a novel kind of augmentation, \augment~ relying on non-systematic scanning of images to alleviate the limitations of ViT.
Specifically, we define four algorithms named Random Patches (RP), Random Tracing (RT), Salient Patches (SP), and Salient Tracing (ST).
Our augmentation, as illustrated in \Figref{method}, 
scans the image and extract patches either,

\noindent
(a) from random scanning of image (low-biased scanning), i.e., Random Patches (RP) or Random Tracing  (RT), or

\noindent
(b) derived with a saliency-guided mechanism (high-biased scanning), i.e., Salient Patches (SP) or Salient Tracing  (ST).

Additionally, it is worth noting that our augmentation approach primarily emphasizes scanning the image.
As a result, it seamlessly integrates with any existing image transformations or augmentations that may be employed before implementing our dynamic scanning technique.
This flexibility allows for incorporating other preprocessing steps or enhancements before applying our dynamic scanning algorithm.

\noindent
\textbf{Our Contributions:}
\begin{description}[style=sameline, nolistsep, leftmargin=*, itemsep=8pt, font=$\bullet~$] %

    \item[Novel augmentation based on dynamic input sequences:]
    Our \augment~ relies on non-systematic scanning of images, creating dynamic input sequences for the transformers.
    Thus, when the transformer processes a single image multiple times, it sees the image differently due to the dynamic input sequence of patches.
    Our augmentation allows us to shuffle the sequence of patches fed to the transformers.
    It also allows the patches to be different between multiple scans of the same image.
    This enables the transformers to focus on different parts of the image between multiple scans, essentially learning more about the image and improving the accuracy of the transformer.

    \item[Adaptive attention:]
    Our investigations reveal that transformers employed with \augment~ attend to different patches/regions depending on the input sequence provided to the transformer.
    This can be seen as the adaptability of attention to dynamic input sequences, suggesting that transformers employed with \augment~ can attend to different regions in an image to induce correct classification.

    \item[Steep robustness against adversarial attacks:]
    We discover that by merely employing \augment~ in transformers, we can significantly boost robustness (around $75\%$) against various adversarial attacks.
    Moreover, this boost comes without degrading the performance of transformers.
    Tests reveal that the crucial reason for improved robustness in transformers is a more adaptive kind of attention enabled by \augment.

\end{description}

\section{Related Works}

Transformers were first proposed in \cite{vaswani2017attention} and showed significant improvement in machine translation performance.
Since the success of transformers and self-attention modules in NLP, researchers have tried to use the transformers in various vision tasks, like
image classification \cite{dosovitskiy2020image,chen2020generative},
object-detection \cite{carion2020end,zheng2020end,dai2021up,zhu2021deformable,sun2021rethinking},
segmentation \cite{chen2021pre,wang2021end,wang2021max,zheng2021rethinking},
image enhancement \cite{chen2021pre,yang2020learning},
image generation \cite{parmar2018image},
human behavior recognition \cite{yang2023efficient},
video in-painting \cite{zeng2020learning}, and
video captioning \cite{zhou2018end}.

ViT, based on pure transformer architecture, achieved comparable performance to the state-of-the-art on multiple image classification benchmarks \cite{dosovitskiy2020image}.
ViT scans the images into a series of non-overlapping patches systematically.
This scanning of images makes learning local features hard since every patch and pixel is processed only once in one context.

However, the performance is still inferior to CNNs when trained from scratch for small and mid-sized datasets.
This drop in performance is attributed by \cite{dosovitskiy2020image} to the lack of inductive biases, such as locality, which are inherent to CNNs.
However, studies from \cite{lu2016dominant,lu2019enhanced,kannan2014salient} shows that the patch-matching algorithm can also be invariant to rotation solving a particular inductive bias.
This suggests that performance in transformers can be further improved with better feature matchings.

Research from \cite{han2021transformer,yuan2021tokens} suggests that ViT inherently learns a poor representation of local features due to this lack of inductive biases.
Many approaches improve ViT by focusing on local features to extract and learn more about the image patches.
One example is Transformer-in-Transformer (TNT) \cite{han2021transformer} architecture, in which an extra transformer module is used to process pixel-level embeddings.
The authors suggest that ViT models the global relation among patch embeddings, while pixel-level embeddings are required to extract local structures.
Another example is Token-to-Token ViT (T2T-ViT) \cite{yuan2021tokens}, where it is suggested that the complex splits and direct tokenizations hinder the learning of local features such as edges.
Research by \cite{zhao2023no} observes that better generalization ability arises from the various distinct features extracted from the input.
Another research from \cite{pang2023caver} suggests that transformers have the capability to interweave fusion structures by rethinking the transformer architecture to include global information about alignment and transformation.

Interestingly, in both the Pyramid Vision Transformer (PVT) proposed by \cite{wang2021pyramid} and the Swin Transformer proposed by \cite{liu2021swin}, the researchers suggest that the Vision Transformer can benefit from a hierarchical (pyramid) structure similar to CNNs to predict finer feature-maps and perform better at pixel-level predictions.
They suggest that ViT's columnar structure, which processes patches at a fixed scale, limits the learning of the transformers.
At the same time, in Swin Transformers, the authors also propose to use the shifted windows approach for finer processing of patches locally, which also allows learning cross-window relationships between non-overlapping patches.

From a different perspective, permutation in input sequence was studied in Attention Neuron proposed by \cite{tang2021sensory} for reinforcement learning tasks.
In this approach, it was observed that when transformers are fed with a permuted input sequence, they exhibit better robustness and generalization.
This article investigates dynamic input sequences created by dynamic scanning of images from a different perspective.
This includes different permutations of input patches and extends the possibility of including a single patch either more than once or not including it in the input sequence.

\section{Overview of \augment~}

\begin{figure*}[!t]
    \centering
    \includegraphics[width=\textwidth]{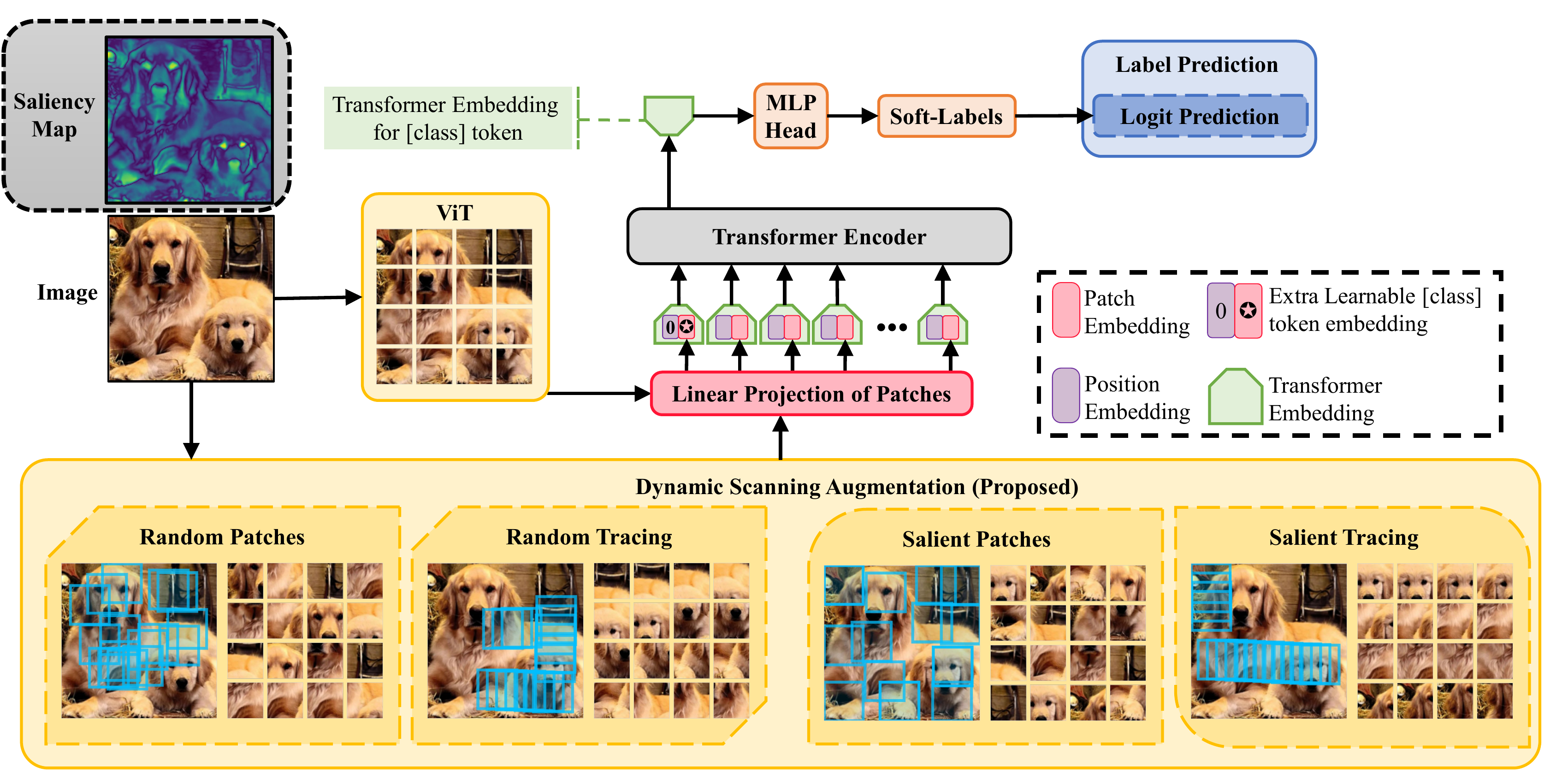}
    \caption{
        Illustration of how the proposed \augment~ differs from ViT.
        The figure also shows how the sequence of patches extracted differs between different variations of \augment.
    }
    \label{overview}
\end{figure*}

In the case of ViT, every patch of the image is only processed once systematically by the transformer, as illustrated in \Figref{overview}.
Moreover, as there are no overlapping pixels and patches, we can infer that transformers only look at the patches and pixels once, limiting the information learned about a patch.
At the same time, the input sequence of patches also remains consistent, i.e., there is no shuffling of patches within the input sequence.

However, transformers are most efficient for NLP Tasks, where every word generally appears more than once in different contexts, making the transformers learn more about the word and the contexts as a whole.
Theoretically, each pixel can also be viewed with eight different contexts based on preceding and succeeding neighboring pixels.
However, viewing every pixel in every context is computationally heavy and often unnecessary.

Therefore, we propose \augment~ to learn more about the image.
This approach extracts the patches dynamically from the image, enabling the transformers to perceive a single image differently when presented multiple times.
Moreover, there is no guarantee that a particular patch will be fed to the transformers.
Transformers must adapt to classify the image based on the given input sequence of the patches.

Below, we give a brief description of the different variations of \augment~ and also note the bias included in scanning using these variants,

\begin{figure*}[!t]
    \centering
    \includegraphics[width=\textwidth]{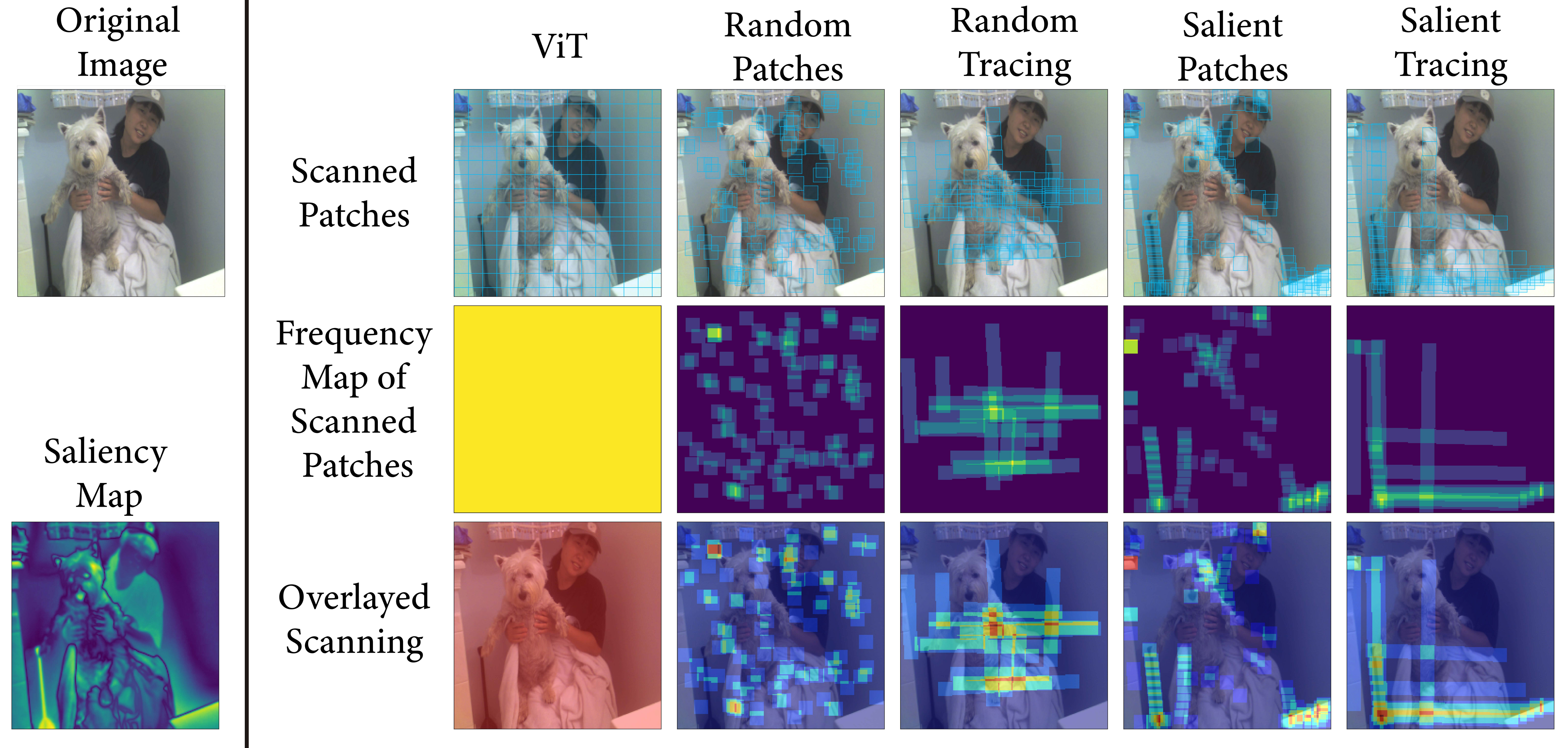}
    \caption{
        Here, we show an example of regions scanned by various \augment~ variants on an image.
        We also show the saliency map of the original image, which is used as a guide for salient patches and salient tracing.
        Further, we plot the frequency of scanned patches.
        This map is homogeneous for ViT since every patch is extracted and processed only once.
        However, for \augment, both possibilities exist: either a patch is extracted multiple times in a single scan or not extracted.
        Therefore, the darker color represents no extraction of patches, while the lighter color represents patches extracted more than once.
        We also overlay this frequency map onto the original image to visualize the regions of the image extracted by our augmentation.
    }
    \label{scan}
\end{figure*}

\begin{algorithm}[!t]
    \caption{Algorithm for extraction of patches using Random Patches}
    \label{algo:random_patches}
    \begin{algorithmic}[1]
        \Require
        \Statex image: The input image
        \Statex num\_patches: The number of patches to extract
        \Statex patch\_size: The size of each patch
        \Ensure
        \Statex patches: The extracted patches
        \Procedure{RandomPatches}{}
        \State patches $\gets$ empty list
        \For{$i \gets 1$ to num\_patches}
        \State coordinate $\gets$ randomly selects a pixel location in the image
        \State patch $\gets$ crop the image using coordinate and patch\_size
        \State add patch to patches
        \EndFor
        \State sequence $\gets$ concatenate all patches in patches
        \State \Return sequence
        \EndProcedure
    \end{algorithmic}
    \textbf{Time Complexity:} $O(\text{num\_patches})$
\end{algorithm}

\begin{algorithm}[!t]
    \caption{Algorithm for extraction of patches using Random Tracing}
    \label{algo:random_tracing}
    \begin{algorithmic}[1]
        \Require
        \Statex image: The input image
        \Statex num\_patches: The number of patches to extract
        \Statex patch\_size: The size of each patch
        \Ensure
        \Statex patches: The extracted patches
        \Procedure{RandomTracing}{}
        \State patches $\gets$ empty list
        \While{length of patches $<$ num\_patches}
        \State selected\_coordinates $\gets$ randomly select two-pixel locations from the image
        \State ray\_direction $\gets$ direction from selected\_coordinates[1] to selected\_coordinates[2]
        \State current\_coordinate $\gets$ selected\_coordinates[1]
        \State patch $\gets$ crop the image using current\_coordinate and patch\_size
        \State add patch to patches
        \While{current\_coordinate $ \neq $ selected\_coordinate[2]}
        \State next\_coordinate $\gets$ find next pixel location along ray\_direction
        \State current\_coordinate $\gets$ next\_coordinate
        \State patch $\gets$ crop the image using current\_coordinate and patch\_size
        \State add patch to patches
        \EndWhile
        \EndWhile
        \State sequence $\gets$ concatenate all patches in patches
        \State \Return sequence
        \EndProcedure
    \end{algorithmic}
    \textbf{Time Complexity:} $O(\text{num\_patches} \cdot \text{path\_length})$, where path\_length represents the average length of the traced paths.
\end{algorithm}

\begin{algorithm}[!t]
    \caption{Algorithm for extraction of patches using Saliency Patches}
    \label{algo:saliency_patches}
    \begin{algorithmic}[1]
        \Require
        \Statex image: The input image
        \Statex saliency\_map: The saliency map of the image
        \Statex num\_patches: The number of patches to extract
        \Statex patch\_size: The size of each patch
        \Ensure
        \Statex patches: The extracted patches
        \Procedure{SalientPatches}{}
        \State patches $\gets$ empty list
        \State pixel\_order $\gets$ sort pixel locations based on saliency\_map in descending order
        \For{$i \gets 1$ to num\_patches}
        \State patch $\gets$ crop the image using pixel\_order[i] and patch\_size
        \State add patch to patches
        \EndFor
        \State sequence $\gets$ concatenate all patches in patches
        \State \Return sequence
        \EndProcedure
    \end{algorithmic}
    \textbf{Time Complexity:} $O(\text{num\_patches})$
\end{algorithm}

\begin{algorithm}[!t]
    \caption{Algorithm for extraction of patches using Salient Tracing}
    \label{algo:salient_tracing}
    \begin{algorithmic}[1]
        \Require
        \Statex image: The input image
        \Statex saliency\_map: The saliency map of the image
        \Statex num\_patches: The number of patches to extract
        \Statex patch\_size: The size of each patch
        \Ensure
        \Statex patches: The extracted patches
        \Procedure{SalientTracing}{}
        \State patches $\gets$ empty list
        \State pixel\_order $\gets$ sort pixel locations based on saliency\_map in descending order
        \State count $\gets 1$
        \While{length of patches $<$ num\_patches}
        \State selected\_coordinates $\gets$ [pixel\_order[count], pixel\_order[count+1]]
        \State ray\_direction $\gets$ direction from selected\_coordinates[1] to selected\_coordinates[2]
        \State current\_coordinate $\gets$ selected\_coordinates[1]
        \State patch $\gets$ crop the image using current\_coordinate and patch\_size
        \State add patch to patches

        \While{current\_coordinate $\neq$ selected\_coordinate[2]}
        \State next\_coordinate $\gets$ find next pixel location along ray\_direction
        \State current\_coordinate $\gets$ next\_coordinate
        \State patch $\gets$ crop the image using current\_coordinate and patch\_size
        \State add patch to patches
        \EndWhile
        \State count $\gets$ count $+ 1$
        \EndWhile
        \State sequence $\gets$ concatenate all patches in patches
        \State \Return sequence
        \EndProcedure
    \end{algorithmic}
    \textbf{Time Complexity:} $O(\text{num\_patches} \cdot \text{path\_length})$, where path\_length represents the average length of the traced paths.
\end{algorithm}

\begin{description}[leftmargin=*]

    \item[Random Patches (RP) (Least Biased):]
    In this variant, we extract the patches randomly (uniform distribution) from all over an image and feed them to transformers in a sequence as defined in Algorithm \ref{algo:random_patches}.
    Since all the patches from an image are extracted randomly, we include minimal bias in scanning the image.

    \item[Random Tracing (RT) (Low Biased):]
    In this variant, we choose two random patches in an image and trace all the patches in the imaginary ray connecting these two chosen patches.
    We do this by tracing through rays repeatedly till we have enough patches for the transformers as defined in Algorithm \ref{algo:random_tracing}.
    We include the bias of the sub-sequence of patches linked to each other since we trace the patches between two random patches.

    \item[Salient Patches (SP) (High Biased):]
    In this variant, we first start by creating a visual guidance map of the image using a saliency map as proposed by \cite{montabone2010human}.
    We then feed the patches to the transformer in order of the saliency of patches as defined in Algorithm \ref{algo:saliency_patches}.
    Further, the saliency map by \cite{montabone2010human} can be referred to as an algorithm to separate the background from the foreground without using any deep learning and relying solely on image processing.
    Hence, this extraction of patches using a guide (saliency map) adds a built-in bias in the scanning where the patches extracted depend on the saliency computed by the algorithm.

    \item[Salient Tracing (ST) (High Biased):]
    In this variant, similar to Salient Patches, we first compute the saliency map of the image as proposed by \cite{montabone2010human} to generate our visual-guidance map.
    Then, similar to Random Tracing, we then trace all the patches that lie in the imaginary rays connecting in-between the salient patches tracked in order of their saliency.
    We do this tracing repeatedly till we have enough patches for the transformers as defined in Algorithm \ref{algo:salient_tracing}.
    Like Salient Patches and Random Tracing, this extraction of patches using a guide also adds an inherent bias to the scanning. It adds extra bias where the sub-sequence of patches are linked to each other.

\end{description}

\Figref{scan} illustrates the different variations of \augment.
We show the scanned patches and plot the frequency map to highlight where a patch is seen multiple times in a scan.
The figure shows that scanning for Random Patches results in extracting patches that are not dependent on any factor within the image itself.
However, as mentioned above, other variations of \augment~ include some bias to extract some patches.

Our \augment~ has the following characteristics;

\noindent
(a) Non-Systematic scanning of images,

\noindent
(b) Stochastic scanning of images,

\noindent
(c) Possibility to re-look at the patches,

\noindent
(d) Hold back some of the available visual information, and

\noindent
(e) Acquiring information in different contexts.

\noindent
We believe that because of these properties, it is possible to
learn more aggressively about robust features, which is verified by the experiments.

As we use non-systematic scanning, we modify the positional embedding used by ViT.
For ViT, the patches' position refers to the position of the patch in the sequence.
Since ViT systematically scans the image, the position of the patch in the sequence also corresponds to the position of the patch in the image, and there exists a fixed number of patches for the transformer.

However, we use the index of the patch's center pixel as the patch's position for our augmentations, as the position of the patch in the sequence is not related to the position of the patch in the image.
Also, the number of patches fed to the \augment~ transformer is controlled with a parameter.
Therefore, if a patch center lies in $(x,y)$ pixel, we convert it to one-dimensional using row-major form as $p = (x \times n) + y$, where $n$ is the image size.
Further, to accommodate the position of the [\textit{class}] token, we shift the raw position of the patches by $1$.
It is to be noted that this position encoding was also evaluated by \cite{dosovitskiy2020image} and showed no change in performance.

In the realm of deep learning and natural language processing, the transformer architecture has a complexity of \(O(N^2)\) where \(N\) represents the sequence input to the transformer \cite{vaswani2017attention}.
ViT's systematic scanning utilizes a fixed value for \(N\), often computed as $ N = (\text{Image Size} / \text{Patch Size})^2$ \cite{dosovitskiy2020image}.
In contrast, our proposed \augment~ allows flexibility to the user for choosing \(N\) to adapt the algorithm's computational resources to the specific requirements of their tasks. Thus, converting \(N\) as a hyperparameter, subject to variation.

\section{Experimental Results}

\subsection{Experimental Design and Settings}

\begin{table}[!t]
    \centering
    \caption{Description of the datasets evaluated in our article.
        We evaluate a small-image size dataset, CIFAR-10, and one big image-size dataset, Imagenette.
        Imagenette is a subset of the bigger ImageNet dataset consisting of only $10$ classes of the original ImageNet dataset.
    }
        \begin{tabular}{l c c}
            \toprule
            \textbf{Dataset}
            & \textbf{CIFAR-10}
            & \textbf{Imagenette}
            \\
            \midrule
            \textbf{Image Size}    & $32  \times 32  \times 3$ & $220 \times 220 \times 3$ \\ 
            \textbf{\# Classes}    & 10     & 10    \\ 
            \textbf{Training Data} & 50,000 & 9,469 \\ 
            \textbf{Testing Data}  & 10,000 & 3,925 \\ 
            \bottomrule
        \end{tabular}

    \label{dataset}
\end{table}

\noindent
\textbf{Datasets:}
We experiment with a small-image size dataset CIFAR-10 \cite{krizhevsky2009learning},
and one big-image size dataset Imagenette \cite{imagenette} (A subset of ILSVRC 2012 ImageNet \cite{russakovsky2015imagenet} with $10$ classes),
to evaluate the performance and embedding of the transformers.
\Tableref{dataset} gives an overview of the datasets

We use Auto-augment \cite{cubuk2018autoaugment} preprocessing to augment the images before training.
Auto-augment modifies the image with several transformations depending on the dataset.
Note that our \augment~ can be complemented with other image manipulation-based augmentations, like auto-augment and preprocessing-based adversarial defenses, to boost performance further.

\noindent
\textbf{Transformer Variant Used:}
We use the \textit{`Base'} variant of the transformers, with $12$ number of transformer layers and self-attention heads, latent vector size $768$, and MLP size $3072$, similar to ViT \cite{dosovitskiy2020image}.
We train all the models with AdamW optimizer with default hyper-parameters, with an initial learning rate of $0.00008$.
We also use a linear decay of learning rate over $100$ epochs with a warmup of $2$ epochs.
We use the HuggingFace library \cite{wolf-etal-2020-transformers} to implement the transformers.

\noindent
\textbf{Adversarial Attacks:}
To test the robustness of various transformers, we test the transformers on different black-box adversarial attacks,

\noindent
(a) Pixel Attack, an $L_0$ norm black-box attack \cite{su2019one},

\noindent
(b) Threshold Attack, an $L_\infty$ norm black-box attack \cite{kotyan2022adversarial}, and

\noindent
(c) HopSkipJump Attack, a query efficient decision-based black box attack with $L_2$ and $L_\infty$ norms variants \cite{chen2020hopskipjumpattack}.

We implement the attacks using the Adversarial Robustness Toolbox (ART) \cite{art2018} library.
We use the default hyper-parameters for the attacks and set the threshold $th$ to $10$ for both Pixel and Threshold attacks.

\textbf{Computation Power Used:}
All the experiments were conducted on a single system with
Intel\textsuperscript{\tiny\textregistered} Core\textsuperscript{\tiny\texttrademark} i9-10900K CPU @ 3.70GHz with 10 cores, 62.7GiB of RAM and a single NVIDIA\textsuperscript{\tiny\textregistered} GeForce\textsuperscript{\tiny\textregistered} RTX 3090.

\begin{figure*}[!t]
    \centering
    \textbf{Amount of Image Covered with Patch Size of 3} \\
    \text{$\ceil{{Image Size} / {Patch Size}}^2$ = 121} \\
    \vspace{0.1cm}
    \hrule
    \vspace{0.1cm}
    \includegraphics[width=\textwidth]{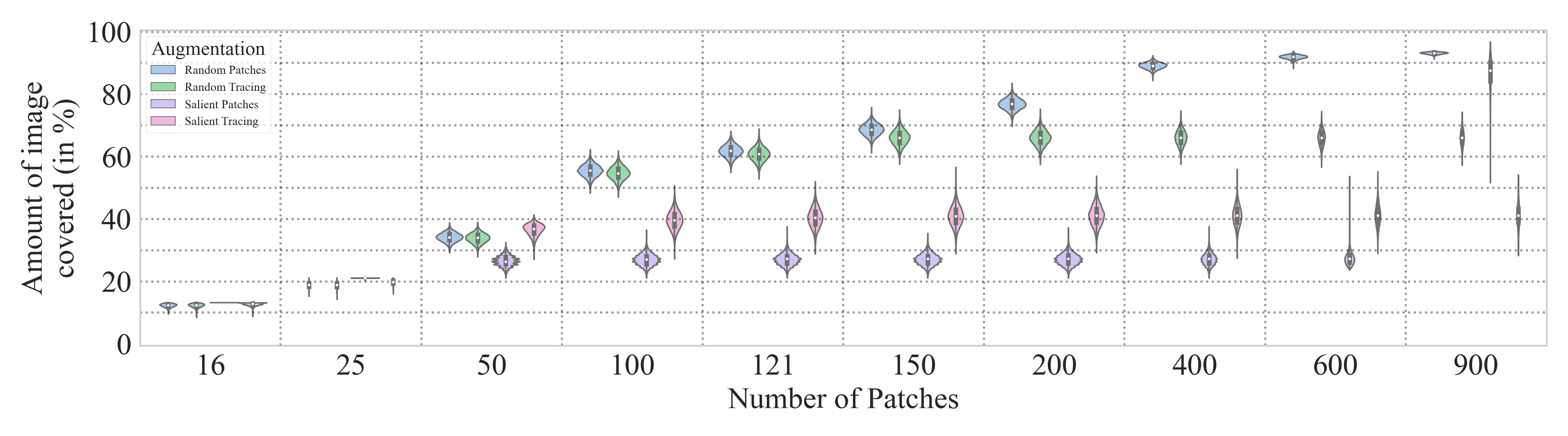} \\
    \vspace{0.2cm}

    \textbf{Amount of Image Covered with Patch Size of 9} \\
    \text{$\ceil{{Image Size} / {Patch Size}}^2$ = 16} \\
    \vspace{0.1cm}
    \hrule
    \vspace{0.1cm}
    \includegraphics[width=\textwidth]{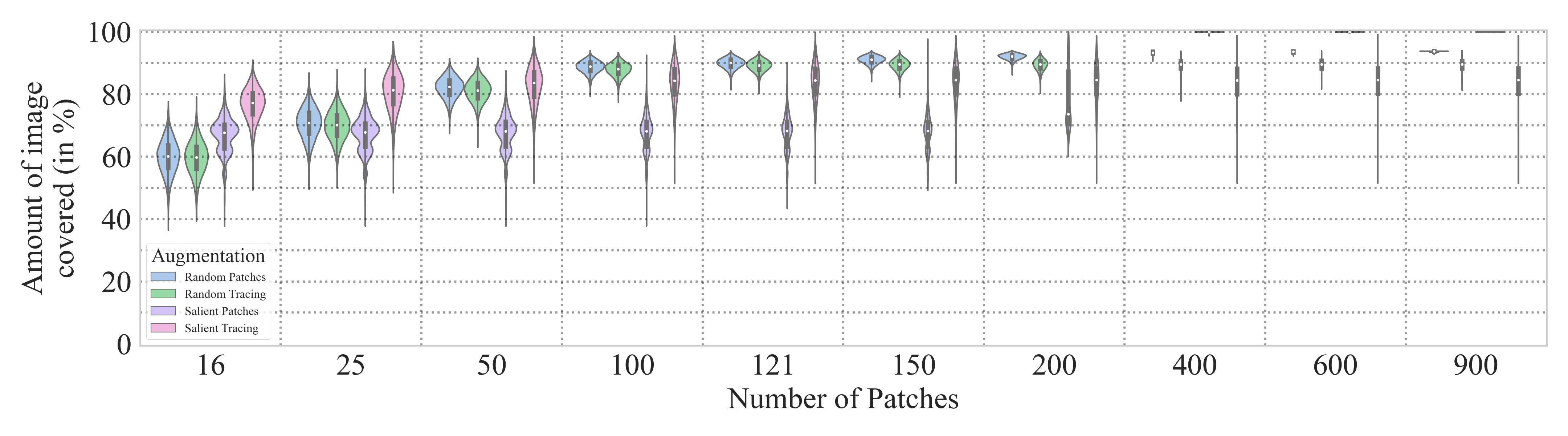} \\
    \vspace{0.1cm}
    \hrule
    \hrule
    \vspace{0.1cm}

    \caption{
        Plot of the amount of the image covered by $N$ patches using different \augment~ for CIFAR-10.
    }
\label{line_tracing_cifar}
\end{figure*}

\begin{figure*}[!t]
\centering
\textbf{Amount of Image Covered with Patch Size of 9} \\
\text{$\ceil{{Image Size} / {Patch Size}}^2$ = 625} \\
\vspace{0.1cm}
\hrule
\vspace{0.1cm}
\includegraphics[width=\textwidth]{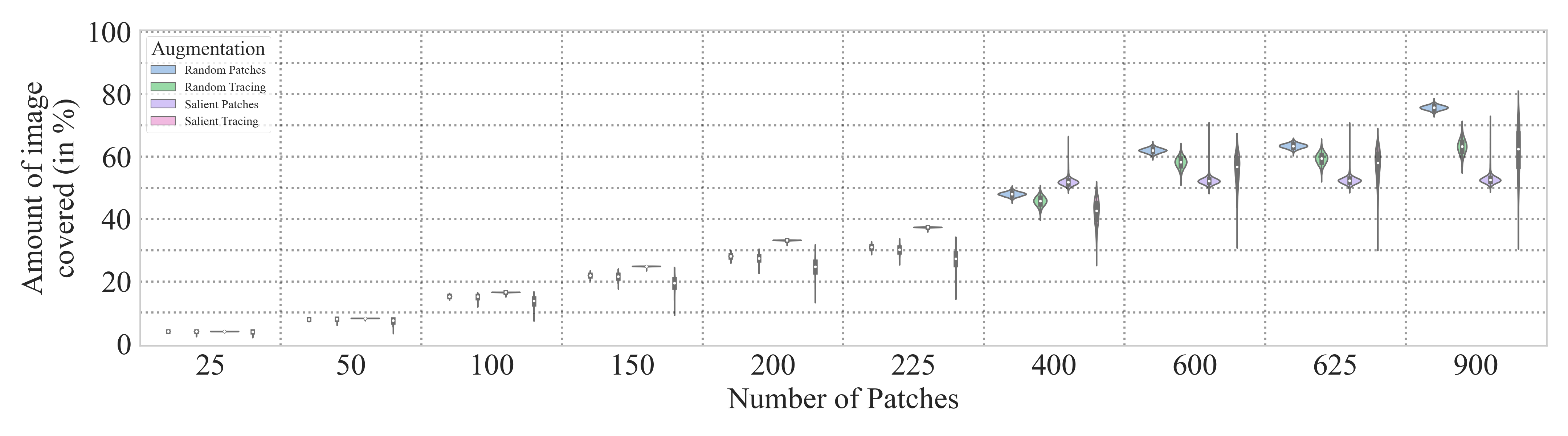} \\
\vspace{0.2cm}

\textbf{Amount of Image Covered with Patch Size of 15} \\
\text{$\ceil{{Image Size} / {Patch Size}}^2$ = 225} \\

\vspace{0.1cm}
\hrule
\vspace{0.1cm}
\includegraphics[width=\textwidth]{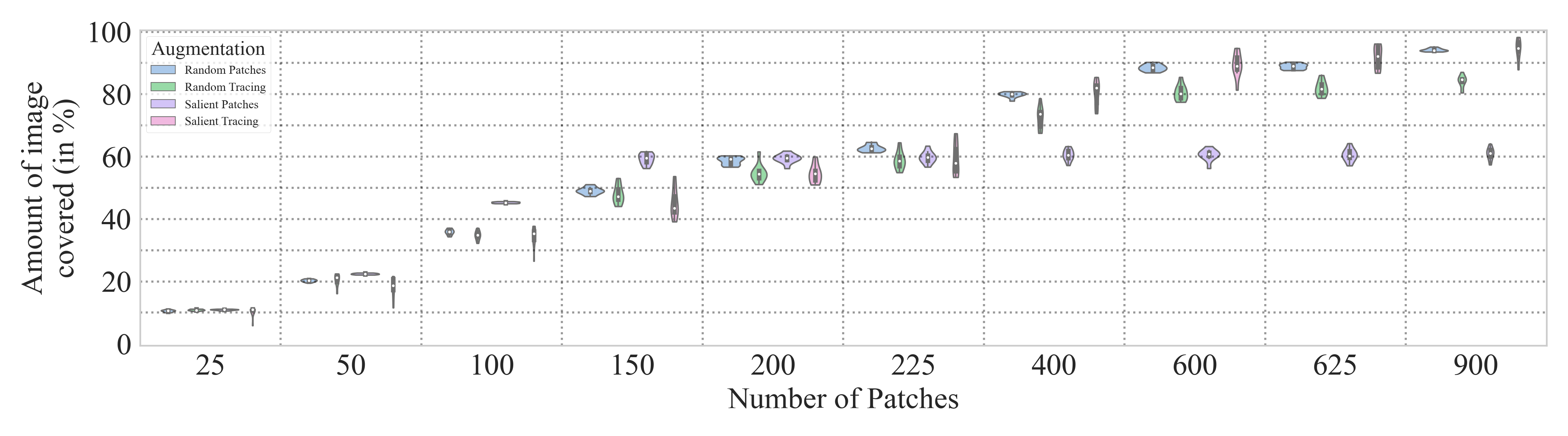} \\
\caption{
    Plot of the amount of the image covered by $N$ patches using different \augment~ for Imagenette.}
\label{line_tracing_imagenet}
\end{figure*}

\textbf{Ablation Tests (Amount of Visual Information):}
An ablation study is conducted to understand the amount of distinct pixels extracted from the image covered by the different number of patches as shown in \Twofigref{line_tracing_cifar}{line_tracing_imagenet}.
Ablation study suggests that \augment~ extract around $55\%-60\%$ of the image on average when using the same number of patches as ViT, that is, $\ceil{\text{Image Size} / \text{Patch Size}}^2$.
The ablation study also shows that the image covered by scans is highly variable in the Salient Patches and Salient Tracing compared to the random counterparts.
This can be accounted for by the bias included in the scanning using a saliency map.
We also notice that the amount of distinct pixels for salient patches remains lower for the Imagenette dataset, suggesting that the objects usually do not cover the entire image, unlike CIFAR-10.
This is expected since, in Imagenette, the object is usually centered and usually does not occupy the entire image.
Further, as \augment~ is non-systematic and stochastic, we report a mean of results over $5$ in different test runs for all the experiments.

\subsection{Performance of \augment~ on small-size dataset}

\begin{table*}[!t]
\centering
\caption{
Accuracy of transformers trained with the different number of patches for the CIFAR-10 dataset.
\textbf{Bold} results correspond to the model having better accuracy than ViT (\Tableref{result_cifar}) for the experiment.
(*) corresponds to the number of patches, which are the same as ViT.
}
\resizebox{\textwidth}{!}{

\begin{tabular}{l r cc cc}
    \toprule
    \textbf{ViT}
    & \textbf{\# of Patches}
    & \textbf{Random Patches}
    & \textbf{Random Tracing}
    & \textbf{Salient Patches}
    & \textbf{Salient Tracing}
    \\
    \midrule
    \multicolumn{6}{c}{\textbf{Patch Size = 3}}  \\
    \midrule \midrule
    & \textbf{25} & 70.73\% $\pm$ 00.18\%          & 57.22\% $\pm$ 00.36\%          & 64.38\% $\pm$ 00.15\% & 54.29\% $\pm$ 00.12\%          \\
    & \textbf{50} & 79.74\% $\pm$ 00.12\%          & 67.96\% $\pm$ 00.21\%          & 75.59\% $\pm$ 00.12\% & 65.67\% $\pm$ 00.17\%          \\
    & \textbf{100} & \textbf{85.08\% $\pm$ 00.15\%} & 78.93\% $\pm$ 00.27\%          & 79.59\% $\pm$ 00.04\% & 76.01\% $\pm$ 00.18\%          \\
    \midrule
    84.15\% & \textbf{*121} & \textbf{85.23\% $\pm$ 00.24\%} & 82.99\% $\pm$ 00.13\%          & 80.91\% $\pm$ 00.06\% & 80.01\% $\pm$ 00.09\%          \\
    \midrule

    & \textbf{150} & \textbf{85.88\% $\pm$ 00.19\%} & \textbf{84.14\% $\pm$ 00.09\%} & 80.58\% $\pm$ 00.11\% & 82.09\% $\pm$ 00.11\%          \\

    & \textbf{250} & \textbf{87.44\% $\pm$ 00.24\%} & \textbf{86.56\% $\pm$ 00.18\%} & 81.77\% $\pm$ 00.14\% & \textbf{85.72\% $\pm$ 00.07\%} \\

    & \textbf{400} & \textbf{88.63\% $\pm$ 00.21\%} & \textbf{88.01\% $\pm$ 00.12\%} & 81.71\% $\pm$ 00.08\% & \textbf{87.72\% $\pm$ 00.11\%} \\

    & \textbf{600} & \textbf{88.91\% $\pm$ 00.09\%} & \textbf{88.32\% $\pm$ 00.16\%} & 83.35\% $\pm$ 00.09\% & \textbf{87.87\% $\pm$ 00.19\%} \\

    & \textbf{900} & \textbf{89.05\% $\pm$ 00.09\%} & \textbf{88.85\% $\pm$ 00.10\%} & 83.37\% $\pm$ 00.06\% & \textbf{89.06\% $\pm$ 00.04\%} \\

    \midrule
    \multicolumn{6}{c}{\textbf{Patch Size = 9}}  \\
    \midrule \midrule

    73.90\% & \textbf{*16} & \textbf{78.17\% $\pm$ 00.22\%} & 61.88\% $\pm$ 00.33\%          & 73.34\% $\pm$ 00.05\%          & 57.42\% $\pm$ 00.35\%          \\
    \midrule

    & \textbf{25} & \textbf{81.91\% $\pm$ 00.15\%} & 66.03\% $\pm$ 00.39\%          & \textbf{74.48\% $\pm$ 00.05\%} & 64.90\% $\pm$ 00.08\%          \\

    & \textbf{50} & \textbf{83.95\% $\pm$ 00.20\%} & \textbf{78.52\% $\pm$ 00.17\%} & \textbf{76.49\% $\pm$ 00.14\%} & \textbf{75.23\% $\pm$ 00.12\%} \\

    & \textbf{100} & \textbf{85.53\% $\pm$ 00.21\%} & \textbf{82.08\% $\pm$ 00.26\%} & \textbf{77.95\% $\pm$ 00.07\%} & \textbf{82.05\% $\pm$ 00.14\%} \\

    & \textbf{150} & \textbf{85.31\% $\pm$ 00.21\%} & \textbf{84.26\% $\pm$ 00.25\%} & \textbf{78.58\% $\pm$ 00.09\%} & \textbf{83.66\% $\pm$ 00.10\%} \\

    & \textbf{250} & \textbf{84.62\% $\pm$ 00.22\%} & \textbf{84.42\% $\pm$ 00.18\%} & \textbf{80.09\% $\pm$ 00.08\%} & \textbf{84.83\% $\pm$ 00.13\%} \\

    & \textbf{400} & \textbf{86.10\% $\pm$ 00.18\%} & \textbf{84.73\% $\pm$ 00.23\%} & \textbf{79.60\% $\pm$ 00.08\%} & \textbf{84.68\% $\pm$ 00.10\%} \\

    & \textbf{600} & \textbf{85.25\% $\pm$ 00.09\%} & \textbf{83.90\% $\pm$ 00.16\%} & \textbf{80.39\% $\pm$ 00.08\%} & \textbf{84.73\% $\pm$ 00.06\%} \\

    & \textbf{900} & \textbf{84.98\% $\pm$ 00.21\%} & \textbf{85.39\% $\pm$ 00.08\%} & \textbf{81.07\% $\pm$ 00.06\%} & \textbf{85.31\% $\pm$ 00.08\%} \\

    \bottomrule
\end{tabular}
}

\label{result_cifar}
\end{table*}

To analyze the performance of ViT and our \augment~ for the small-sized dataset, we first report accuracy over the test dataset of CIFAR-10, where the transformers were trained on the training dataset (\Tableref{result_cifar}).
To the best of our knowledge, this is the first time CIFAR-10's result will be reported for the ViT-Base model trained on CIFAR-10 from scratch.
Results show that the performance of \augment~ using Random Patches is superior to ViT and other variants of augmentation when the number of patches of the image is the same as ViT.

The experimental results show that the variations of \augment~ relying on saliency maps (Salient Patches and Salient Tracing) perform poorly compared to the random variants.
We suspect this happens due to the bias in the extraction of patches, as the salient variants focus only on the most salient parts (presumed foreground) of images.
We also notice that variations of dynamic augmentation relying on tracing (Random Tracing and Salient Tracing) performed poorly for bigger patch sizes (for the same number of patches as ViT), suggesting that bias introduced by tracing where the sub-sequence of image patches depends on each other is not scalable to bigger image sizes.

Further, to verify that visual information by providing distinct pixels in the input sequence has an impact on performance, we experiment with transformers trained with a different number of patches (\Tableref{result_cifar}).
This ensures that we feed the transformers either with more or less distinct patches, manipulating the image's visual information.
We observe an increase in accuracy as the number of patches increases for all the variations of \augment, suggesting that more visual information benefits the transformers.
Here also, the results show the superiority of Random Patches over other variations of \augment.
We also notice an increase in patch size for salient patches, and salient tracing results in superior performance over smaller patch sizes compared to ViT, suggesting that our \augment~ is more suitable for bigger patch sizes.

However, the performance increases substantially for \augment~ relying on tracing when the number of patches increases, suggesting that different context information is necessary to improve the performance on the CIFAR-10 dataset where there are limited pixels.

\subsection{Performance of \augment~ on large-size dataset}

\begin{table*}[!t]
\centering
\caption{
Accuracy of transformers trained with the different number of patches for the Imagenette dataset.
\textbf{Bold} results correspond to the model having better accuracy than ViT (\Tableref{result_imagenette}).
(*) corresponds to the number of patches that are the same as ViT.
}
\resizebox{\textwidth}{!}{

\begin{tabular}{l r cc cc}
    \toprule

    \textbf{ViT}
    & \textbf{\# of Patches}
    & \textbf{Random Patches}
    & \textbf{Random Tracing}
    & \textbf{Salient Patches}
    & \textbf{Salient Tracing}
    \\
    \midrule
    \multicolumn{6}{c}{\textbf{Patch Size = 9}}  \\
    \midrule \midrule

    & \textbf{25}   & 58.14\% $\pm$ 00.14\%          & 27.56\% $\pm$ 00.47\%          & 41.36\% $\pm$ 00.22\%          & 20.58\% $\pm$ 00.42\% \\

    & \textbf{50}   & 66.36\% $\pm$ 00.26\%          & 32.54\% $\pm$ 00.37\%          & 47.85\% $\pm$ 00.13\%          & 26.74\% $\pm$ 00.23\% \\

    & \textbf{100}   & 71.32\% $\pm$ 00.20\%          & 44.93\% $\pm$ 00.55\%          & 54.69\% $\pm$ 00.13\%          & 40.32\% $\pm$ 00.32\% \\

    & \textbf{150}   & 72.39\% $\pm$ 00.34\%          & 54.59\% $\pm$ 00.69\%          & 59.59\% $\pm$ 00.17\%          & 43.84\% $\pm$ 00.34\% \\

    & \textbf{250}   & 73.34\% $\pm$ 00.63\%          & 63.72\% $\pm$ 00.58\%          & 68.74\% $\pm$ 00.15\%          & 48.66\% $\pm$ 00.36\% \\

    & \textbf{400}   & 74.94\% $\pm$ 00.29\%          & 70.56\% $\pm$ 00.20\%          & 74.98\% $\pm$ 00.26\%          & 50.48\% $\pm$ 00.15\% \\

    & \textbf{600}   & \textbf{77.10\% $\pm$ 00.29\%} & 74.06\% $\pm$ 00.52\%          & 76.36\% $\pm$ 00.31\%          & 53.68\% $\pm$ 00.41\% \\

    \midrule
    75.97\% & \textbf{*625}   & \textbf{77.40\% $\pm$ 00.27\%} & 73.90\% $\pm$ 00.47\%          & 75.73\% $\pm$ 00.24\%          & 51.58\% $\pm$ 00.36\% \\
    \midrule

    & \textbf{900}   & \textbf{77.37\% $\pm$ 00.28\%} & \textbf{77.15\% $\pm$ 00.49\%} & \textbf{75.90\% $\pm$ 00.19\%} & 56.38\% $\pm$ 00.31\% \\

    \midrule
    \multicolumn{6}{c}{\textbf{Patch Size = 15}} \\
    \midrule

    & \textbf{25}  & 62.74\% $\pm$ 00.39\%          & 29.21\% $\pm$ 00.69\%          & 45.66\% $\pm$ 00.12\%          & 17.91\% $\pm$ 00.56\% \\

    & \textbf{50}  & 68.98\% $\pm$ 00.21\%          & 35.09\% $\pm$ 00.52\%          & 52.48\% $\pm$ 00.08\%          & 29.24\% $\pm$ 00.29\% \\

    & \textbf{100}  & \textbf{73.00\% $\pm$ 00.61\%} & 49.68\% $\pm$ 00.40\%          & 62.67\% $\pm$ 00.22\%          & 43.08\% $\pm$ 00.15\% \\

    & \textbf{150}  & \textbf{75.07\% $\pm$ 00.32\%} & 55.72\% $\pm$ 00.54\%          & 68.34\% $\pm$ 00.14\%          & 46.37\% $\pm$ 00.25\% \\

    \midrule
    71.87\% & \textbf{*225}  & \textbf{74.98\% $\pm$ 00.23\%} & 63.44\% $\pm$ 00.55\%          & 70.35\% $\pm$ 00.17\%          & 49.93\% $\pm$ 00.35\% \\
    \midrule

    & \textbf{250}  & \textbf{73.71\% $\pm$ 00.24\%} & 64.50\% $\pm$ 00.41\%          & 70.61\% $\pm$ 00.27\%          & 50.86\% $\pm$ 00.53\% \\

    & \textbf{400}  & \textbf{75.25\% $\pm$ 00.37\%} & \textbf{71.07\% $\pm$ 00.61\%} & \textbf{72.62\% $\pm$ 00.28\%} & 53.14\% $\pm$ 00.23\% \\

    & \textbf{600}  & \textbf{77.54\% $\pm$ 00.38\%} & \textbf{75.66\% $\pm$ 00.22\%} & \textbf{73.62\% $\pm$ 00.28\%} & 58.08\% $\pm$ 00.29\% \\

    & \textbf{900}  & \textbf{78.21\% $\pm$ 00.22\%} & \textbf{75.33\% $\pm$ 00.46\%} & \textbf{74.05\% $\pm$ 00.07\%} & 59.67\% $\pm$ 00.33\% \\

    \bottomrule
\end{tabular}
}

\label{result_imagenette}
\end{table*}

To analyze the performance of ViT and our \augment~ for the large-size dataset, we first report accuracy over the test dataset of Imagenette, where the transformers were trained on the training dataset (\Tableref{result_imagenette}).
Results show that the performance of \augment~ using Random Patches is again superior to ViT and other augmentation counterparts when the number of patches is the same as ViT.

Interestingly, we observe some steep drop in performance for the salient tracing variant of \augment.
Since salient tracing extracts extra information between the salient patches and good performance for salient patches, the information available between the salient patches hinders classification.

Next, we also experiment with transformers trained with the different number of patches (\Tableref{result_imagenette}).
We again observe an increase in accuracy as the number of patches increases for all the variants of \augment.
Here also, the results show the superiority of Random Patches over other variants of \augment.

Results from both small-sized (\Tableref{result_cifar}) and large-sized datasets (\Tableref{result_imagenette}) suggest that our \augment~ can benefit from more distinct visual information fed to the transformers trained.
This performance of \augment~ also suggests that not all information (patches) present in the image is required for processing.
Also, our \augment~ performance does not degrade for a bigger patch size.
At the same time, it is reported that ViT suffers from performance degradation as the patch size increases \cite{dosovitskiy2020image}.
The experiments support our hypothesis of learning local features as there is no significant drop in performance on increasing the patch size.

\begin{figure*}[!t]
\centering
\includegraphics[width=\textwidth]{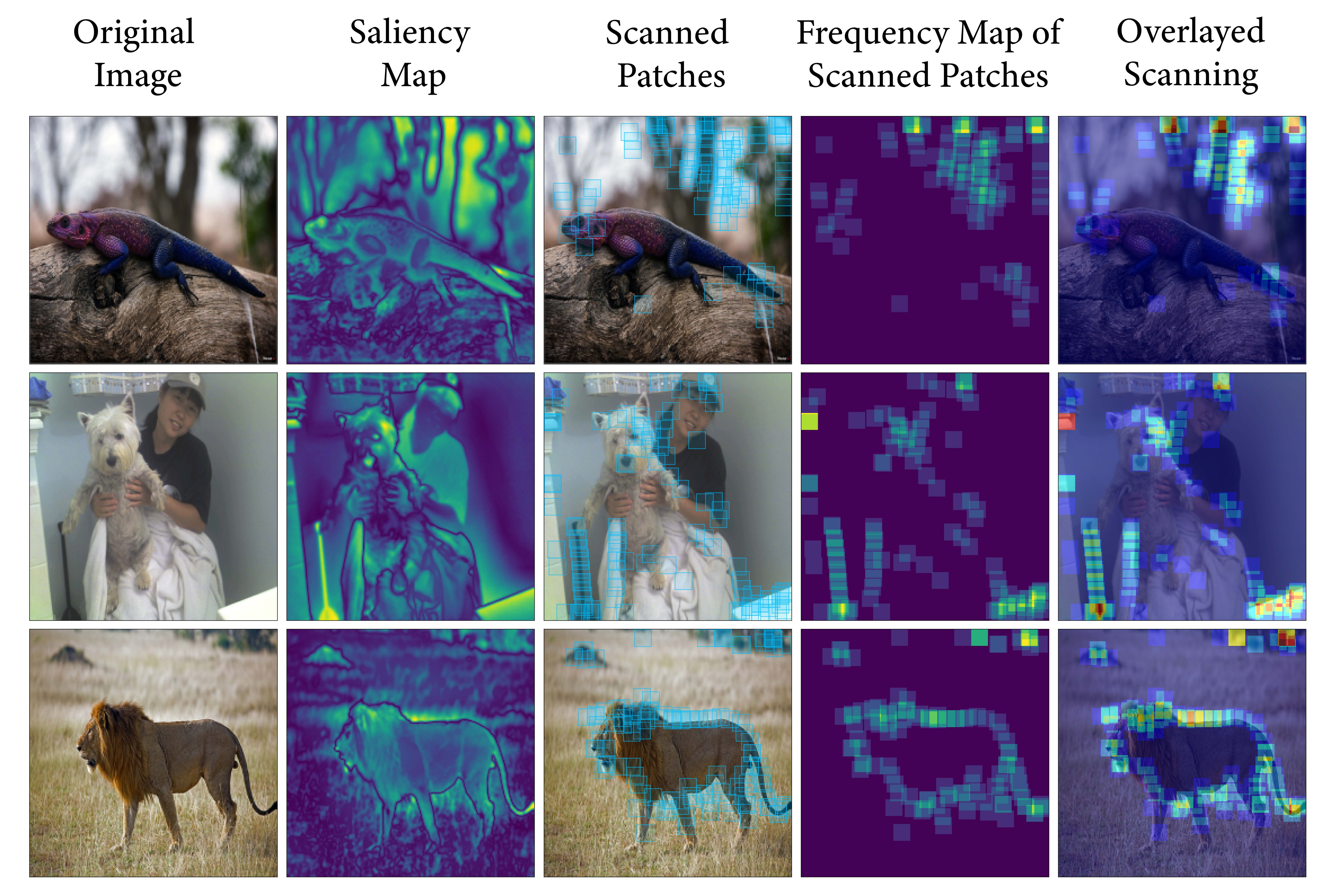}
\caption{
Regions of images scanned by Salient Patches variant of \augment.
We plot, a) Saliency Map, b) Scanned Patches, c) Frequency Map of Scanned Patches, and d) An Overlay of Frequency Map on the original image.
}
\label{saliency_bias}
\end{figure*}

\subsection{Robustness of \augment~ against Black-Box Adversarial Attacks}

\begin{table*}[!t]
\centering
\caption{
Robust Accuracy for different adversarial attacks over random $100$ images for models trained with CIFAR-10 with patch size $3$.
We test Pixel Attack (PA), Threshold Attack (TA), and two variants of Hop Skip Jump Attack (HSJA).
We also report a mean robustness across the four attacks.
Value in parentheses refers to the accuracy of transformers for the random $100$ images.
\textbf{Bold} results in each column correspond to the model having the best robustness against the adversarial attack.
}
\resizebox{\textwidth}{!}{
\begin{tabular}{l c cc cc}
    \toprule
    \multirow{2}{*}{\textbf{Attacks}}
    & \multirow{2}{*}{\textbf{ViT}}
    & \textbf{Random} 
& \textbf{Random} 
& \textbf{Salient} 
& \textbf{Salient} 
\\
&
& \textbf{Patches} 
& \textbf{Tracing} 
& \textbf{Patches} 
& \textbf{Tracing} 
\\
\midrule
\multicolumn{6}{c}{\textbf{\# of Patches = 121 (Same as ViT) }} \\
\midrule \midrule
\textbf{PA ($L_0$)}              & 26.74\% (86\%) & \textbf{90.24\% (84\%)} & 83.53\% (85\%) & 24.42\% (87\%) & 48.78\% (81\%) \\
\textbf{TA ($L_\infty$)}     & 00.00\% (86\%) & \textbf{92.86\% (84\%)} & 86.05\% (85\%) & 35.23\% (87\%) & 56.25\% (81\%) \\
\textbf{HSJA ($L_\infty$)} & 32.56\% (86\%) & \textbf{90.48\% (84\%)} & 80.00\% (85\%) & 21.84\% (87\%) & 48.78\% (81\%) \\
\textbf{HSJA($L_2$)}      & 09.30\% (86\%) & \textbf{84.88\% (84\%)} & 77.64\% (85\%) & 24.42\% (87\%) & 51.85\% (81\%) \\
\midrule
\textbf{Mean} & 17.15\% (86\%) & \textbf{89.62\% (84\%)} & 81.80\% (85\%) & 26.48\% (87\%) & 51.42\% (81\%)\\
\midrule
\multicolumn{6}{c}{\textbf{\# of Patches = 900}} \\
\midrule \midrule
\textbf{PA ($L_0$)}              & 26.74\% (86\%) & 84.62\% (90\%)          & \textbf{86.90\% (86\%)} & 28.41\% (88\%) & 63.33\% (89\%) \\
\textbf{TA ($L_\infty$)}     & 00.00\% (86\%) & 96.63\% (90\%)          & \textbf{97.70\% (86\%)} & 48.28\% (88\%) & 79.31\% (89\%) \\
\textbf{HSJA ($L_\infty$)} & 32.56\% (86\%) & \textbf{92.31\% (90\%)} & 86.05\% (86\%)          & 19.32\% (88\%) & 61.62\% (89\%)         \\
\textbf{HSJA ($L_2$)}      & 09.30\% (86\%) & \textbf{94.44\% (90\%)} & 92.94\% (86\%)          & 18.18\% (88\%) & 73.03\% (89\%)  \\
\midrule
\textbf{Mean} & 17.15\% (86\%) & \textbf{92.00\% (90\%)} & 90.89\% (86\%) & 32.00\% (88\%) & 69.32\% (89\%) \\
\bottomrule
\end{tabular}
}

\label{adv_result}
\end{table*}

\begin{figure*}[!t]
\centering
\includegraphics[width=\textwidth]{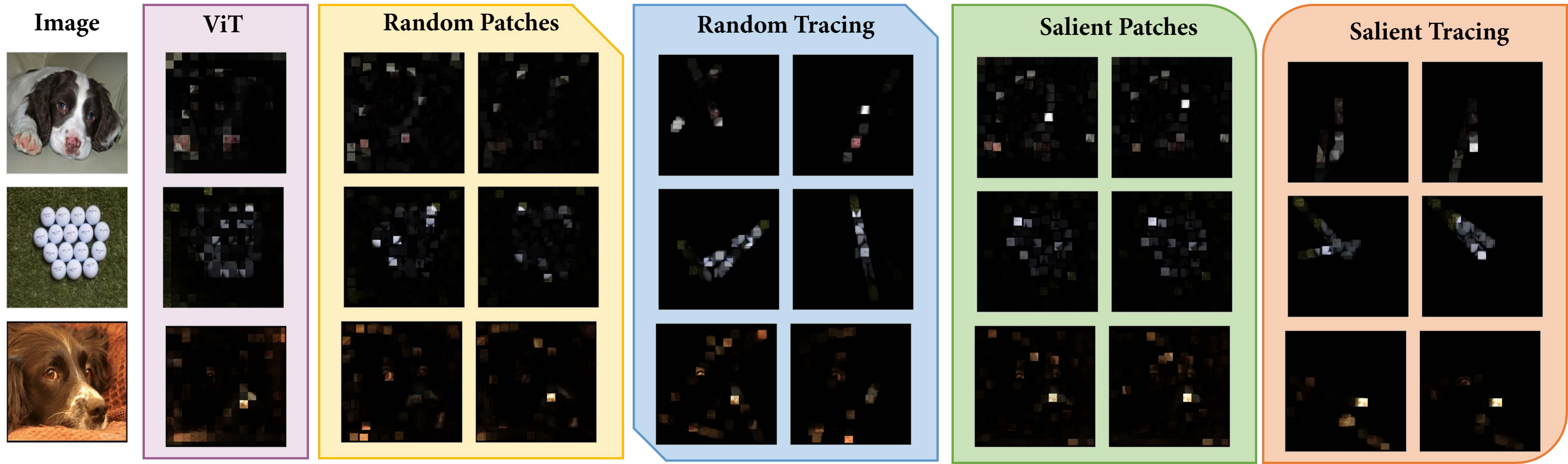}
\caption{
Patches of different images attended by ViT and variations of \augment.
Due to the non-systematic scanning, we show patches attended using two different extractions for our variations of \augment.
This shows that changing the input sequence of patches for the same image corresponds to a change in attended patches for \augment transformers.
}
\label{attention}
\end{figure*}

\begin{table*}[!t]
\centering
\caption{
Robust accuracy of random variants of \augment~ evaluated with stochastic characteristic (when multiple scans of the same image are different) and non-stochastic characteristic (when multiple scans of the same image are alike).
Value in parentheses refers to the accuracy of transformers for the random $100$ images.
\textbf{Bold} results in each column correspond to the model having the best robustness against the adversarial attack.
}
\resizebox{\textwidth}{!}{
\begin{tabular}{l c cc cc}
\toprule
\multirow{3}{*}{\textbf{Attacks}}
& \multirow{3}{*}{\textbf{ViT}}
& \multicolumn{2}{c}{\textbf{Stochastic}}
& \multicolumn{2}{c}{\textbf{Non-Stochastic}}
\\
&
& \textbf{Random} 
& \textbf{Random} 
& \textbf{Random} 
& \textbf{Random} 
\\
&
& \textbf{Patches} 
& \textbf{Tracing} 
& \textbf{Patches} 
& \textbf{Tracing} 
\\
\midrule
\textbf{PA}     & 26.74\% (86\%) & \textbf{90.24\% (84\%)} & 83.53\% (85\%) & 82.09\% (85\%) & 83.86\% (82\%) \\
\textbf{TA}     & 00.00\% (86\%) & \textbf{92.86\% (84\%)} & 86.05\% (85\%) & 92.26\% (85\%) & 94.64\% (82\%) \\

\bottomrule
\end{tabular}
}

\label{adv_result_sto}
\end{table*}

To analyze the performance of ViT and our \augment~ against different black-box adversarial attacks, we report the robustness of the transformers on the test set of the CIFAR-10 dataset.
Specifically, we test against,
(a) Pixel Attack \cite{su2019one},
(b) Threshold Attack \cite{kotyan2022adversarial}, and
(c) Hop Skip Jump Attack \cite{chen2020hopskipjumpattack}.
We attack the random $100$ images correctly classified by the transformers.

\Tableref{adv_result} shows that evaluated adversarial attacks perform terribly over transformer trained with random variants of \augment~ and perform poorly over transformer trained with salient variants of \augment.
Concurrently, the same attacks are quite successful against ViT.
While the ViT can only correctly classify around $17\%$ of attacked images, transformers trained with random patches can correctly classify around $90\%$ of attacked images.
This suggests that our \augment~ has higher resiliency against black-box norm-based adversarial attacks.

Further, as salient variants are more biased than random variants, the experiment suggests that reducing the bias in the scanning of images leads to higher resiliency against adversarial attacks.
Interestingly, despite having superior performance over non-attacked samples, Salient patches are more vulnerable to adversarial attacks than Salient Tracing.
This suggests that when models focus on only salient regions of the images, it is easier for the adversarial attacks to find adversarial perturbations to induce misclassification.

\section{Discussion and Analysis}

\subsection{On the Effect of Scanning Bias on Performance of Transformers}

From Tables \ref{result_cifar}-\ref{result_imagenette}, we notice a substantial difference in the performance of salient variants and random variants of \augment.
On closer inspection, we noticed that the saliency map generated by the algorithm proposed by \cite{montabone2010human} often failed to assign high saliency to objects of interest in the image.
We show three examples in \Figref{saliency_bias}, in which a prominent salient region is identified in the background in the first row. At the same time, in the second row, we notice that the salient region is diffused in both objects of interest (dog) and the background. In contrast, the significant salient region consists of the object of interest (lion) in the last row.
This incorrect variance detection of object-of-interest limits the performance of the salient variants of the \augment as patches containing information about the object of interest are often not fed to transformers.

Since the scanning is biased to extract salient patches from the image in the salient variants, there is often a chance that the information about the object of interest is insufficient to classify.
Moreover, as we increase the number of patches, we increase the chances of the patches containing information about the object of interest being fed to the transformer. Therefore, we notice an increase in performance.
In contrast, the random variants extract patches from all over the image and generally have better performance.

\subsection{Investigation of Adaptive Attention}

Due to the systematic scanning of images, ViT can focus on the input sequence of patches in only one way.
This creates a one-to-one correspondence between the input space (images) and the output space (classes).
However, our investigation shows that transformers can learn to focus on different patches in multiple ways depending on different input sequences of patches by using \augment.
This creates a dynamic many-to-one correspondence between the input space (images) and output space (classes).
Traditional augmentation practices modify images to have similar many-to-one correspondence.

Our augmentation relies on scanning the image by transformers rather than modifying the image to find multiple correspondences from input space to output.
Therefore, \augment~ networks have attention to images that can adapt depending on the input sequences of patches.
This adaptability in attention can be visualized as shown in \Figref{attention}, where for \augment, there exist multiple attended patches for the same image but different input sequences.

Since our augmentations enable the transformers to have adaptability in attention, this induces adaptability in classification that naturally increases the robustness of the network.
Since adversarial perturbation affects a patch or group of patches, the model can still try to focus on other patches to give the correct classification.
This is also supported by our experiments using adversarial attacks (\Tableref{adv_result}).
Interestingly, we also found that variations of \augment~ relying on saliency (biased scanning) were less rigorous in finding multiple ways to classify, which have poor robustness than the random variants that explored multiple ways more rigorously.

In order to effectively evaluate the robustness and adaptability in classification, we also attack the transformers trained by keeping the scan for an image fixed for adversarial attacks, which usually processes the images more than once to optimize perturbations (\Tableref{adv_result_sto}).
Experimental robustness reveals that stochastic patch extraction (different input sequences in multiple scans of the same image) plays a minor role in elevating resiliency against adversarial attacks.
However, the crucial contributor to the adversarial robustness of \augment~ transformers lies in the adaptability in the image classification.

\section{Conclusion}

This article proposes various types of \augment~ related to each other based on dynamic scanning of images and dynamic input sequences to the transformers.
We show that the transformers augmented with dynamic scanning augmentation are more accurate and robust than the ViT.
Transformers trained with Random Patches outperform the ViT in all datasets tested with over $5\%$ improvement in standard accuracy and over $75\%$ increase in robustness across various adversarial attacks.
Our investigations also show that injecting bias into the transformers lead to a degradation of performance.
As Random Patches, the variant with the least bias outperforms all other variants of \augment.
In fact, our \augment~ does not use all available pixels; on average, only $60\%$ of the image pixels is enough to achieve superior results than ViT.
Moreover, increasing the distinct pixels in the input sequence also boosts the performance of the transformers trained with our augmentation.
A significant contributing factor to robustness is adaptability in attention due to the dynamic input sequence of patches.
Our experiments reveal that transformers employed with \augment~ attend to different patches/regions depending on the input sequence provided to the transformer.
We define this adaptability in attention as the capability to attend to different patches/regions depending on different images' scans (input sequences of patches), which plays a vital role in contributing to resiliency against adversarial attacks.
Thus, we show that non-systematic scanning of images improves the performance and robustness of standard Vision Transformers.

\section*{Acknowledgments}

This work was supported by JSPS Grant-in-Aid for Challenging Exploratory Research - Grant Number JP22K19814, JST Strategic Basic Research Promotion Program (AIP Accelerated Research) - Grant Number JP22584686, JSPS Research on Academic Transformation Areas (A) - Grant Number JP22H05194.

\com{

    \com{
        \textbf{Human Vision:}
        Humans explore the world with a series of eye movements \cite{kandel2000principles}.
        We explore our visual surroundings in a series of quick saccades (glimpses), which move the fovea (a part of the retina) from one fixation point to another, which results in a new image for the brain with every saccade \cite{becker1989neurobiology}.
        Moreover, we perceive a stable visual surrounding even though the image on the fovea shifts with each saccade \cite{wurtz2008neuronal}.
        This scanning by saccades is triggered endogenously to explore the visual surrounding.
        We also have selective visual attention as we focus more on areas of interest, and background objects are mostly ignored \cite{yarbus1967eye, henderson1999high}.

        While both humans and transformers rely on processing input sequences, gazing at humans to scan the surroundings is fundamentally different from how the ViT systematically scans the image.
        Further, gazing in humans acquire information in multiple contexts, unlike ViT, which only processes each patch once.
        To improve the learning for transformers, we are motivated to scan the image non-systematically. 
        %
    }

    \com{
    \section{Limitations and Societal Impact}

    A potential limitation of our augmentation is vulnerability against adversarial attacks.
    Current adversarial attacks are static-optimization-based algorithms that try to find adversarial perturbations while the input and output to the networks are constant.
    Due to \augment~ transformers' inherent adaptive nature, \augment~ converts the problem of adversarial machine learning into dynamic optimization, which is more challenging than static optimization but not unsolvable.
    Hence, \augment~ remains vulnerable to future adversarial attacks based on dynamic optimization.
    A societal impact of the article lies in the number of patches processed by the transformers.
    A direct consequence of increasing the number of patches is the requirement for more
    GPU memory compared to ViT to process the patches effectively.
    }



}

\bibliographystyle{iclr2024_conference}
\bibliography{adversarial_machine_learning}

\end{document}